# Neural Language Models with Distant Supervision to Identify Major Depressive Disorder from Clinical Notes

Bhavani Singh Agnikula Kshatriya, MS[1], Nicolas A Nunez, MD[3], Manuel Gardea-Resendez, MD[3], Euijung Ryu, PhD[2], Brandon J Coombes, PhD[2], Sunyang Fu, MHI[1], Mark A Frye, MD[3], Joanna M Biernacka, PhD[2,3], Yanshan Wang, PhD[1]

[1]Department of Artificial Intelligence & Informatics, Mayo Clinic, Rochester, MN; [2]Department of Quantitative Health Sciences, Mayo Clinic, Rochester, Minnesota USA; [3]Department of Psychiatry & Psychology, Mayo Clinic, Rochester, Minnesota USA

**Abstract**

Major depressive disorder (MDD) is a prevalent psychiatric disorder that is associated with significant healthcare burden worldwide. Phenotyping of MDD can help early diagnosis and consequently may have significant advantages in patient management. In prior research MDD phenotypes have been extracted from structured Electronic Health Records (EHR) or using Electroencephalographic (EEG) data with traditional machine learning models to predict MDD phenotypes. However, MDD phenotypic information is also documented in free-text EHR data, such as clinical notes. While clinical notes may provide more accurate phenotyping information, natural language processing (NLP) algorithms must be developed to abstract such information. Recent advancements in NLP resulted in state-of-the-art neural language models, such as Bidirectional Encoder Representations for Transformers (BERT) model, which is a transformer-based model that can be pre-trained from a corpus of unsupervised text data and then fine-tuned on specific tasks. However, such neural language models have been underutilized in clinical NLP tasks due to the lack of large training datasets. In the literature, researchers have utilized the distant supervision paradigm to train machine learning models on clinical text classification tasks to mitigate the issue of lacking annotated training data. It is still unknown whether the paradigm is effective for neural language models. In this paper, we propose to leverage the neural language models in a distant supervision paradigm to identify MDD phenotypes from clinical notes. The experimental results indicate that our proposed approach is effective in identifying MDD phenotypes and that the Bio-Clinical BERT, a specific BERT model for clinical data, achieved the best performance in comparison with conventional machine learning models.

**Introduction**

Major depressive disorder (MDD) is a prevalent worldwide psychiatric disorder that is associated with high rates of functional impairment and healthcare use[1,2,3]. In the United States it has been estimated that 8% of adults over 20 years old have clinically significant depressive symptoms such as persistent feeling of sadness and loss of interest[4,5]. Major depressive episodes, it can arise for example in the context of a unipolar (only depressive episodes) or bipolar depression (occurrence of mania/hypomania and depressive episodes); major depressive episodes are characterized by a two week period of pervasive symptoms of low mood or anhedonia and five of the of following symptoms: low energy, apathy or agitation, guilt or worthlessness, suicide ideation, poor concentration, psychomotor agitation/inhibition, change in appetitive or weight. The Diagnostic and Statistical Manual of Mental Disorders, 5th Edition (DSM-5) has proposed different specifiers such as anxious distress, atypical features, melancholic features, catatonia, mixed features, seasonal patterns, with psychotic features or of peri-partum onset[6]. Unfortunately, current modifications in modern diagnostic classification systems, such as ICD-9 or ICD-10 are still inadequately to adequately capture the heterogenous MDD phenotypes thus, leading to misdiagnosis and unsuccessful treatment outcomes[7]. Although, depressive episodes can be diagnosed effectively depressive symptoms tend to overlap with other psychiatric disorders (e.g. attention deficit hyperactivity disorder; substance use disorders) creating nosological difficulties and leading to misdiagnosis[7]. For example, different studies have compared illness trajectory between bipolar depression and major depressive disorders underscoring distinguishing clinical features such as presence of atypical symptoms (increased appetite/sleep, psychomotor retardation) and subsyndromal manic symptoms which were mostly associated to BP depressions[8,9,10]. Consequently, this highlights the importance of correctly phenotyping patients to ascertain diagnostic criteria.

There have been many attempts to develop MDD phenotyping algorithms in the literature. The approaches for phenotyping not only differ at algorithmic level but also based on the type of data being used. Electroencephalographic (EEG) data was frequently used in the studies for phenotyping of MDD. Certain studies have generated images from EEG signals while others have extracted non-linear features from EEG data to classify MDD[11,12]. There was an algorithmic diversity using EEG data among various studies involving implementation of machine learning models such as logistic regression, SVM along with other deep learning models including convolutional neural networks, inception, and ResNet models to identify MDD[11,12,13]. MDD classification can also be performed using specific ICD codes apart from EEG data. But related studies have illustrated that using ICD codes alone is unreliable and produce inaccurate phenotypic results[14,15]. Combinatorial approaches have also been employed for MDD phenotyping using data sources such as ICD codes and medications orders, but not incorporating clinical notes[16].

Adoption of Electronic Health Records (EHR) led to significant increase in the availability of valuable patient information specifically hidden in the free text. Extensive natural language processing (NLP) techniques have been employed in the clinical domain to extract data for diverse healthcare applications[17,18]. As a major portion of EHR data, clinical notes have been utilized to ascertain other psychiatric illnesses using various algorithms namely rule-based NLP models, SVM and others[19,20]. These clinical notes contain abundant information which may provide more accurate phenotyping which requires development of NLP algorithms to identify depression related concepts from the notes. While extensive MDD-related phenotypic information is also documented in clinical notes, only a few studies utilized clinical notes for phenotyping MDD[20], and, to the best of our knowledge, no studies have applied deep learning based neural language models for identifying MDD from clinical notes. One of the major reasons for not applying deep learning models in the field of MDD phenotyping using clinical notes might be the lack of large annotated training datasets. Distant supervision is a simple and adaptable approach leveraging programmatically created weakly labeled training sets in training machine learning models. In prior studies, researchers have utilized the distant supervision approach to train machine learning models on the clinical text classification tasks to mitigate the issue of lacking annotated training data in the clinical domain[21,22,23]. In our study, we explore the distant supervision approach to create weakly labeled data to train the MDD phenotyping algorithm.

Deep learning based neural language models have been implemented in the clinical domain to perform various tasks including but not limited to named entity recognition, relation extraction, and predictive modeling. Various methods such as Long Short Term Memory (LSTM), Recurrent Neural Networks (RNN), and others are popular in the clinical domain but they have several limitations including long term contextual dependency issues[24,25]. Transfer learning implementation in NLP domain was a significant breakthrough which resulted in the ability to use pre-trained representations in downstream tasks such as classification[26]. Recent advancements in NLP resulted in state-of-the-art Bidirectional Encoder Representations for Transformers (BERT) model which is a transformer-based model that can pre-train bi-directional language representations from a corpus of unsupervised text data[27]. The BERT model was pretrained on BooksCorpus and English Wikipedia data and thus has significant contextual information embedded in it. It outperformed other models like Elmo in various NLP tasks including classification[27]. BERT and its variants have shown its potential applicability and better performance comparatively in the clinical domain to perform classification/phenotyping tasks[28,29]. Considering the performance and architecture of BERT, researchers have trained and deployed domain specific BERT models such as Bio-Clinical BERT (trained on MIMIC, an openly available dataset based on ~60,000 intensive care health care admission) which was initialized on Bio-BERT has clinical contextual information and thus can perform better in clinical applications compared to the original BERT[30].

To the best of our knowledge, the distant supervision paradigm has not been evaluated for BERT and its variants of neural language models. In this paper, we propose to apply the state-of-the-art neural language models with distant supervision paradigm to identify MDD phenotypes using clinical notes. In addition, we compared our approach with conventional machine learning models namely Random Forest, K-Nearest Neighbors, and Support Vector Machine Classifiers, in which we used Word2Vec[31,32] word embeddings to generate numerical feature vectors, to demonstrate the effectiveness of the proposed approach and the advantage over traditional NLP models for MDD identification.

**Method**

**Data**

In this study, we used the Mayo Clinic Biobank (MCB) cohort[33,34] to build a corpus of clinical note documents. The MCB has more than 56,000 patients who have biological samples, self-reported health information, and with comprehensive EHR data to enable research studies[35]. Using this cohort to develop MDD phenotyping algorithms will enable us to conduct various future studies such as association between genotypes and phenotypes. We first used the MDD-related ICD codes in Table 1 to identify the cases and controls. The case cohort are patients with >=2 MDD-related ICD codes while the control cohort contains patients without any MDD-related ICD codes. For this study, we have used 1000 patients (randomly selected 500 cases and 500 controls) to develop MDD phenotyping algorithms. We then retrieved the consultation clinical notes from the Mayo Clinic Data Warehouse which consisted of 168,139 documents. Two datasets were created using these clinical notes, which include distant supervised/weakly-labeled training clinical documents and 500 gold standard testing clinical documents. The goal is sentence-level identification of MDD instances, i.e., Positive (e.g., "Patient has history of dysthymia"), Possible (e.g., "Patient is a depression suspect"), Negative (e.g., "There is no evidence of depression"), or unknown MDD instances. An annotation guideline was developed to manually annotate a gold standard clinical note dataset for testing purpose.

**Table 1**. MDD-related ICD codes used in this study.

| ICD-9 | ICD-10 |
|---|---|
| 296.2, 296.21, 296.22, 296.23, 296.24, 296.25, 296.26, 296.3, 296.3, 296.31, 296.32, 296.33, 296.34, 296.35, 296.36, 296.82, 298, 298.0, 300.4, 311 | F32, F32.0, F32.1, F32.2, F32.3, F32.4, F32.5, F32.8, F33, F33.0, F33.1, F33.2, F33.3, F33.4, F33.40, F33.41, F33.42, F33.8, F34.1, F33.9 |

**Gold standard test data.** The dataset was generated by randomly selecting 500 documents from the total documents. Two psychiatrists (NAN and MGR) annotated all 500 documents manually using the annotation guidelines for positive, possible, and negated MDD at the sentence level using Multi-Document Annotation Environment (MAE) tool[36]. All the remaining sentences of the document were labeled as unknown sentences using simple regex-based NLP code. The total sentences retrieved resulted in 28,461 sentences. The test set was highly imbalanced with only ~0.98% MDD related sentences.

**Distant supervision.** The weakly labeled training data was generated using the remaining 167,639 clinical documents. A rule-based NLP model was developed using clinician suggested keywords and patterns from annotated sentences as shown in Table 2 and implemented using Medtagger NLP pipeline[37,38]. All the clinical documents were parsed through the pipeline and labeled for either positive, possible or negated MDD (Figure 1). The remaining sentences were labeled as unknown instances. This process generated 3,081,359 sentences. The dataset was highly imbalanced with 98% unknown instances and 2% (i.e., 53,569) MDD related instances. In order to achieve case balance, we have under-sampled the unknown cases to 53,569 sentences (the number of MDD related sentences). Thus, the final train set consists of 51,047 documents with 107,138 sentences. We haven't under-sampled the train data to the level of minority class i.e., Negated MDD as it would result in significant reduction in training data which might lead to poor performance in both machine learning and deep learning models. The train set was further split into 99% train and 1% validation set.

Though the unknown cases were under sampled to the equal number of MDD related cases, there was still significant imbalance among the four classes between training and gold standard test set as shown in Table 3.

**Table 2:** Rule-based NLP algorithm for Identification of MDD

| **Keywords** | depressive disorder major, mdd, major depressive illness, depressive phase, poor motivation, depression, anhedonia, depressive disorder/disorders, depressive symptoms, dysthymia, mood disorder, dysthymic, low mood, seasonal affective disorder |
|---|---|
| **NLP Rules** | \b(anhedonia\|depressive disorder(s)?\|poor motivation\|depression\|dysthymia\|mood disorder\|dysthymic\|low mood\|seasonal affective disorder\|depressive phase\|mdd)(?!\s+Scale\|\s+scale\|\s+equal\s+1)\b |

**Neural Language Models**

We have used the state-of-the-art transformer model known as Bidirectional Encoder Representation for Transformers (BERT) to identify MDD cases. There are two types of BERT used in the study which include 1) BERT[27] and 2) Bio-Clinical BERT[30]. BERT is pretrained on Wikipedia and BookCorpus whereas Bio-Clinical BERT was pretrained using MIMIC-III and initialized on top of Bio BERT. Both models differ in the vocabulary size with BERT having vocabulary size of 30,522 and Bio-Clinical BERT with 28,996. For BERT, we have used a base uncased model considering its less computational requirements.

An additional dropout layer along with a classifier layer with output dimension of four was added on top of pretrained BERT. Cross entropy loss function was used to penalize the incorrect predictions with Adam optimizer. The model was trained with a maximum text sequence length of 128, batch size of 32, and 3e-5 cyclical learning rate for 6 epochs.

Bio-Clinical BERT was implemented using similar architecture and hyperparameter settings. Both models were implemented using Pytorch Huggingface repository[39].

**Baseline Machine Learning Models**

We have implemented three baseline machine learning algorithms to identify MDD from clinical notes, namely Random Forest Classifier (RF), K-Nearest Neighbors (K-NN), and Support Vector Machine (SVM). Similar to the neural language models, these conventional machine learning models are used in the distant supervision paradigm with the same experimental settings. Since machine learning models require text to be transformed into numerical feature vectors independently unlike BERT, we have used word embeddings to generate the feature vector of words as it is one of the vital techniques to not only transform words to vectors but also capture the context information around the words. Word embeddings can be generated using Skip Gram (or) Common Bag of Words (CBOW) method. CBOW uses the contextual representations within a specific window to predict the word. The weights learned by CBOW act as vector representations of text otherwise known as word embeddings. Word2Vec is an algorithm with a combination of these methods. We have used the Gensim library Word2Vec algorithm and trained it from scratch on the train data using CBOW method within a window of 5. The trained model was used to generate the word vectors with a dimension of 300 for both train and test set.

**Figure 1:** Distant Supervision approach to generate weakly labeled training data for identifying MDD

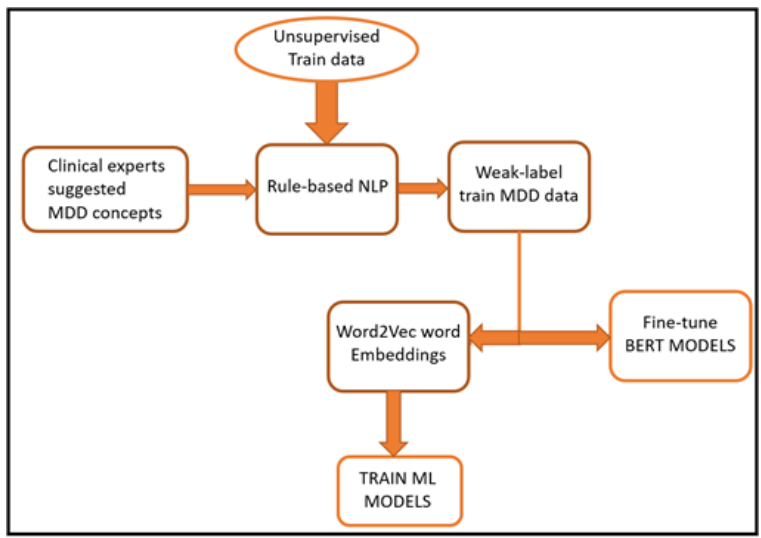

**Table 3:** Data imbalance among various classes

| MDD class | unknown | Positive MDD | Possible MDD | Negative MDD |
|---|---|---|---|---|
| % imbalance (train) | 50% | 44.5% | 3.6% | 1.9% |
| % imbalance (test) | 99% | 0.825% | 0.087% | 0.066% |

The following is the brief description and specific parameter settings for the three machine learning models. RF is an ensemble model with many decision trees generated from subsets of training data. The output class is defined by the majority voting from all the decision trees. A total of 100 estimators were used to train and test our RF model. K-NN model computes the classification of the data point by using the majority vote of the data points in the nearest neighbors. We have used uniform weights for all the categories of data points and K value of 7 through grid search to train the model. SVM tries to compute an optimal hyperplane to divide the data points into their respective classes.

Squared L2 regularization function is applied to penalize the incorrect classification. We have used a linear SVM kernel with parameter C value of 1.

**Evaluation Metrics**

We have evaluated the machine learning and deep learning models using Precision, Recall, and F1 score metrics.

## Results

Table 4 lists the results of MDD identification using various machine learning models and neural language models with distant supervision. The results indicate that the Bio-clinical BERT achieved the best performance among all the models with 0.95 F1 score for positive MDD phenotype identification, 0.87 F1 score for possible MDD phenotype identification, 0.91 F1 score for negated MDD phenotype identification, and 0.99 F1 score for unknown cases. Both BERT variant models have better performance compared to conventional machine learning models. BERT is inferior to Bio-Clinical BERT in terms of identifying Possible and Negated MDD phenotypes. Among three conventional machine learning models, RF performed the best with 0.51 F1 score for Positive MDD phenotype identification, 0.32 F1 score for Possible MDD phenotype identification, 0.40 F1 score for Negated MDD phenotype identification, and 0.99 F1 score for unknown cases. All the tested models achieved good performance in identifying unknown cases due to availability of high amounts of data.

**Table 4:** Comparison of results for MDD identification among various models (**P**= Precision, **R**= Recall, **F1**= F1 score).

| MODEL | unknown | | | Positive MDD | | | Possible MDD | | | Negated MDD | | |
|---|---|---|---|---|---|---|---|---|---|---|---|---|
| | *P* | *R* | *F1* | *P* | *R* | *F1* | *P* | *R* | *F1* | *P* | *R* | *F1* |
| **Bio-clinical BERT** | 0.99 | 0.99 | **0.99** | 0.92 | 0.98 | **0.95** | 0.91 | 0.84 | **0.87** | 1.0 | 0.84 | **0.91** |
| **BERT** | 0.99 | 0.99 | **0.99** | 0.86 | 0.96 | 0.91 | 0.71 | 0.68 | 0.69 | 0.83 | 0.79 | 0.81 |
| **Random Forest** | 0.99 | 0.99 | 0.99 | 0.35 | 0.94 | 0.51 | 0.83 | 0.20 | 0.32 | 0.55 | 0.32 | 0.40 |
| **K-Nearest Neighbors** | 0.99 | 0.90 | 0.94 | 0.08 | 0.93 | 0.15 | 0.04 | 0.44 | 0.07 | 0.07 | 0.58 | 0.12 |
| **SVM classifier** | 1.0 | 0.98 | 0.99 | 0.24 | 0.91 | 0.38 | 0.0 | 0.0 | 0.0 | 0.33 | 0.05 | 0.09 |

Table 5 provides error analysis of MDD identification from Bio-Clinical BERT, BERT, and RF. Bio-Clinical BERT was able to identify MDD instances with critical clinical terms needed to differentiate MDD instances such as anhedonia, dysthymia as shown in Table 5 (example: (2, b); (1, c); (3, d)). BERT was not able to identify such instances as its pretraining contextual representations are not from clinical text like Bio-Clinical BERT. BERT showed good performance in identifying cases which have general grammatical representations without much of clinical context as shown Table 5 (such as example: (4 in d, c); (3, b)).

**Discussion**

In this paper, we have used explicit MDD related concepts to abstract MDD phenotypes. However, it is not uncommon that the MDD related concepts were not explicitly mentioned in the clinical notes. For example, major symptoms related to MDD, such as appetite, sleep, mental agitation, reduced brain function, low mood, sadness, hopelessness, tiredness and extreme behavior, might be mentioned in the clinical notes, which potentially could indicate MDD. Consideration of these symptoms in addition to the explicit MDD concepts results in high confidence diagnosis of

MDD patients. In the future work, we plan to extend our study to implement our distant supervision and deep learning approach to identify those major symptoms.

**Table 5:** Error analysis of MDD identification among various models (**C**=Correct, **NC**=Not Correct).

| **Examples** | **Bio-clinical BERT** | **BERT** | **Random Forest** |
|---|---|---|---|
| a) **Unknown** | | | |
| 1. No seasonality to mood concerns | C | NC | NC |
| 2. The patient mood is dysphoric | C | C | NC |
| 3. Requests evaluation of anxiety and depression for patient | NC | NC | NC |
| 4. Patient main complaint is of fatigue on interview | NC | C | NC |
| 5. There is a strong family history of depression | NC | NC | C |
| b) **Positive MDD** | | | |
| 1. Note of doctor where he completely details her blood pressure readings, hyperlipidemia, and depression | C | C | NC |
| 2. Patient has hx of dysthymia | C | NC | NC |
| 3. Transitioning from Prozac, been on long for depression to Cymbalta for control of musculoskeletal pain | NC | C | NC |
| c) **Possible MDD** | | | |
| 1. Possible OSA; Hx of dysthymia, in remission | C | NC | NC |
| 2. Asked to evaluate the patient for depression and possible transfer | C | C | NC |
| 3. The case is a depression suspect | NC | NC | NC |
| 4. Patient risk of intentionally ending his life in the immediate future is low | NC | C | NC |
| d) **Negated MDD** | | | |
| 1. Test for folate, and vitamin D to rule out any unidentified etiologies for depression | NC | NC | NC |
| 2. There were no frank signs of depression. | C | C | NC |
| 3. Likewise, he is not experiencing anhedonia. | C | NC | NC |
| 4. His PHQ-9 score of zero suggests absence of depression | NC | C | NC |

Though Bio-Clinical BERT performed better than other models, there is a significant space for improvement in its performance in identifying MDD. There are certain limitations in our study, firstly, the data imbalance with low MDD related instances. Cost sensitivity approach can help tackle the data imbalance problem by penalizing higher for the incorrect predictions for minority classes specifically in machine learning algorithms to improve and evaluate their performance. Although data imbalance problems can be handled by BERT due their ability to learn the contextual representations, cost sensitivity needs to be also implemented in the BERT when data distribution is dissimilar. Secondly, the interpretability of the BERT models. We had to perform manual analysis of the results and errors to interpret BERT model's ability to identify correct and incorrect instances comparatively using several examples as it is not possible to comprehend the reason for deep learning models predictions explicitly. Thirdly, clinicians might have overlooked or emphasized certain symptoms than others resulting in selection bias on the keywords used in the study. Fourth, we have used the cohort which contain white population and as of we know there might be differences in longitudinal course illness with other populations. Apart from these limitations, our study has shown the effectiveness of generating weak-label data to classify MDD cases by using NLP rules resulting in reduction of manual effort. The generalizability of the Bio-Clinical BERT on clinical data from other institutions has not been studied in our research but considering the Bio-Clinical BERT ability of having strong contextual knowledge of clinical terminology, it might be easily generalizable compared to other models. There is a potential to improve machine learning models, but it requires greater refinement in generating word embeddings and experimenting with various parameters of the models.

**Conclusion**

MDD is a prevalent psychiatric disorder that has been associated with significant healthcare burden worldwide. Phenotyping MDD could help facilitate early diagnosis and consequently has significant advantages on treatment outcomes. Most of the MDD phenotypic information has been documented in free-text EHR data, such as clinical notes. These clinical notes may provide more accurate phenotyping information, but extracting this information requires development of natural language processing (NLP) algorithms. In this study, we applied the state-of-the-art neural language models with distant supervision paradigm to identify MDD phenotypes using clinical notes. We generated weakly labeled data using a rule-based NLP algorithm for training neural language models. Although the resulting data was highly imbalanced, we were able to label comparatively a greater number of training data to facilitate and improve manual chart review. We evaluated neural language models BERT and Bio-Clinical BERT, and compared with three conventional machine learning models, namely RF, K-NN, and SVM models for the identification of MDD in 500 clinical documents manually annotated by two psychiatrists to test the performance of the proposed models creating a gold standard for the dataset. Our results showed that the BERT variant models achieved better performance compared to three machine learning models. Bio-Clinical BERT outperforms other models in the identification of all three categories of MDD due to its contextual understanding of clinical terminology compared to BERT. This also emphasizes that the domain specific pretraining of BERT models can result in significant performance improvements. The experiment also validates that the distant supervision is an effective approach to generate weakly labeled training data for neural language models to identify efficaciously MDD from clinical notes.